\documentclass[journal]{IEEEtran}

\usepackage{amsmath, amsthm, amssymb, amsfonts}
\usepackage{epsfig,subfigure}
\usepackage{color}
\usepackage{multirow}
\usepackage{siunitx} 
\newcolumntype{C}[1]{>{\centering\arraybackslash}m{#1}}

\def\f{{\mathbf f}}

\def\p{{\mathbf p}}
\def\r{{\mathbf r}}
\def\q{{\mathbf q}}
\def\t{{\mathbf t}}
\def\w{{\mathbf w}}
\def\x{{\mathbf x}}
\def\y{{\mathbf y}}

\def\P{{\mathbf P}}
\def\W{{\mathbf W}}
\def\Q{{\mathbf Q}}
\def\X{{\mathbf X}}
\def\C{{\mathbf C}}

\def\Y{{\mathbf Y}}
\def\lrule{\leaders\hrule height3pt depth-2.6pt\hskip2em \relax}

\newcommand{\AMIN}{\mathop{{\rm argmin}}\limits}

\begin{document}
\title{Rank-1 Convolutional Neural Network}
%
%
%

\author{Hyein~Kim, 
        Jungho~Yoon,
        Byeongseon Jeong,
        and~Sukho~Lee
\thanks{H. Kim, and J. Yoon are with the Department
of Mathematics, Ewha University, Seoul, Korea,}
\thanks{B. Jeong is with the Ewha Institute of Mathematical Sciences, Ewha University, Seoul, Korea,
}
\thanks{and S. Lee is with the Division of Computer Engineering, Dongseo University, Busan, Korea
e-mail: petrasuk@gmail.com.}
\thanks{Manuscript submitted August 11, 2018.}}

%
%
\markboth{Preprint August~2018}%
{Shell \MakeLowercase{\textit{et al.}}: Bare Demo of IEEEtran.cls for IEEE Journals}
%



\maketitle

\begin{abstract}
In this paper, we propose a convolutional neural network(CNN) with 3-D rank-1 filters which are composed by the outer products of 1-D vectors. After being trained, the 3-D rank-1 filters can be decomposed into 1-D filters in the test time for fast inference. The reason that we train 3-D rank-1 filters in the training stage instead of consecutive 1-D filters is that a better gradient flow can be obtained with this setting, which makes the training possible even in the case where the network with consecutive 1-D filters cannot be trained.
The 3-D rank-1 filters are updated by both the gradient flow and the outer product of the 1-D vectors in every epoch, where the gradient flow tries to obtain a solution which minimizes the loss function, while the outer product operation tries to make the parameters of the filter to live on a rank-1 sub-space. 
Furthermore, we show that the convolution with the rank-1 filters results in low rank outputs, constraining the final output of the CNN also to live on a low dimensional subspace.
\end{abstract}


%
\IEEEpeerreviewmaketitle

\section{Introduction}

\begin{figure}
\centering
	\includegraphics[width=0.8\columnwidth]{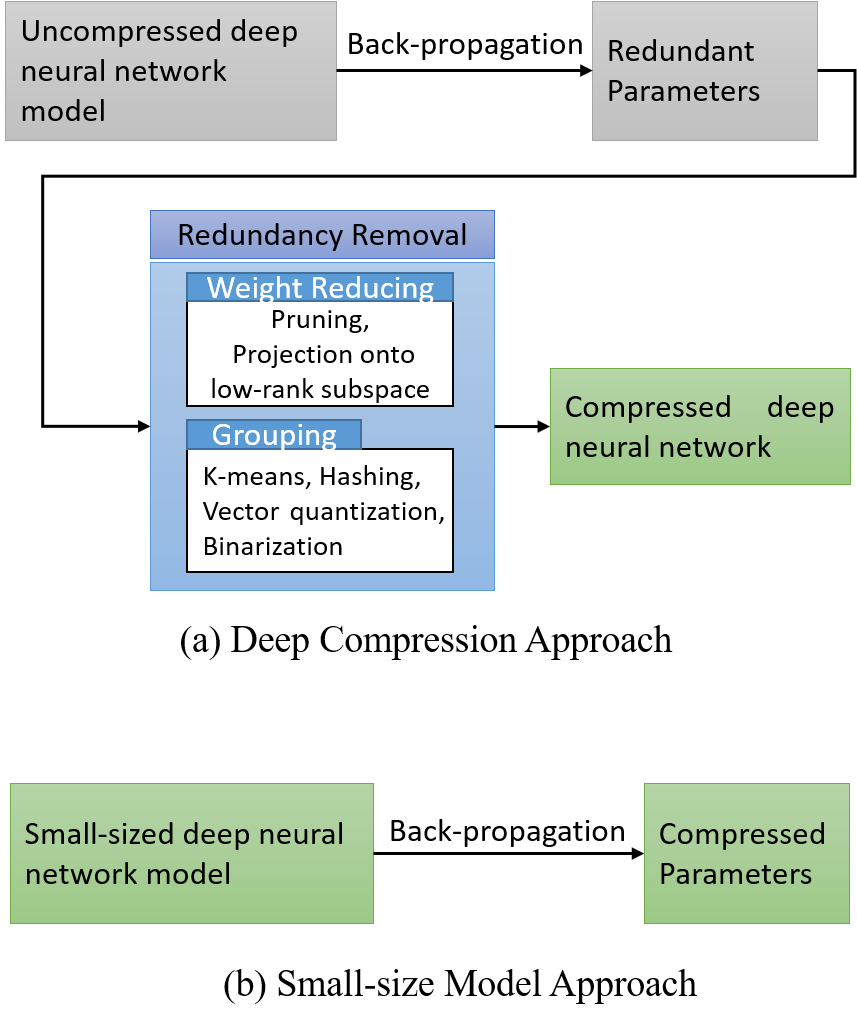} 
    	\caption{Two kinds of approaches trying to achieve small and efficient deep models (a) approach of compressing pre-trained parameters (b) approach of modeling and training a small-sized model directly.}
    \label{frameworks}
\end{figure}

\IEEEPARstart{N}{owdays} deep convolutional neural networks (CNNs) have achieved top results in many difficult image classification
tasks. However, the number of parameters in CNN models is high which limits the use of deep models on devices with limited resources such as smartphones, embedded systems, etc.
Meanwhile, it has been known that there exist a lot of redundancy between the parameters and the feature maps in deep models, i.e., that CNN models are over-parametrized.  
The reason that over-parametrized CNN models are used instead of small sized CNN models is that the over-parametrization makes the training of the network easier as has been shown in the experiments in \cite{Livni}. The reason for this phenomenon is believed to be due to the fact that the gradient flow in networks with many parameters achieves a better trained network than the gradient flow in small networks. 
Therefore, a well-known traditional principle of designing good neural networks is to make a network with a large number of parameters, and then use regularization techniques to avoid over-fitting rather than making a network with small number of parameters from the beginning.\\ 
\indent However, it has been shown in \cite{Zhang} that even with the use of regularization methods, there still exists excessive capacity in the trained networks, which means that 
the redundancy between the parameters is still large. 
This again implies the fact that the parameters or the feature maps can be expressed in a structured subspace with a smaller number of coefficients. 
Finding the underlying structure that exist between the parameters in the CNN models and reducing the redundancy of parameters and feature maps are the topics of the deep compression field.  
As has been well summarized in \cite{CompressDeep1}, researches on the compression of deep models can be categorized into works which try to eliminate unnecessary weight parameters \cite{CompressDeep2}, works which try to compress the parameters by projecting them onto a low rank subspace \cite{CompressDeep3}\cite{CompressDeep4}\cite{CompressDeep5}, and works which try to group similar parameters into groups and represent them by representative features\cite{CompressDeep6}\cite{CompressDeep7}\cite{CompressDeep8}\cite{CompressDeep9}\cite{CompressDeep10}.
These works follow the common framework shown in Fig. \ref{frameworks}(a), i.e., 
they first train the original uncompressed CNN model by back-propagation to obtain the uncompressed parameters, and then try to find a compressed expression for these parameters to construct a new compressed CNN model.\\
\indent In comparison, researches which try to restrict the number of parameters in the first place by proposing small networks are also actively in progress (Fig. \ref{frameworks}(b)). However, as mentioned above, the reduction in the number of parameters changes the gradient flow, so the networks have to be designed carefully to achieve a trained network with good performance. 
For example, MobileNets \cite{Mobilenet} and Xception networks \cite{Xception} use depthwise separable convolution filters, while the Squeezenet \cite{Squeezenet} uses a bottleneck approach to reduce the number of parameters.
Other models use 1-D filters to reduce the size of networks such as the highly factorized Flattened network \cite{Flattened}, or the models in \cite{TrainingLow} where 1-D filters are used together with other filters of different sizes. 
Recently, Google's Inception model has also adopted 1-D filters in version 4.
One difficulty in using 1-D filters is that 1-D filters are not easy to train, and therefore, they are used only partially like in the Google's Inception model, or in the models in \cite{TrainingLow} etc., except for the Flattened network which is constituted of consecutive 1-D filters only. 
However, even the Flattened network uses only three layers of 1-D filters in their experiments, due to the difficulty of training 1-D filters with many layers.\\
 \indent In this paper, we propose a rank-1 CNN, where the rank-1 3-D filters are constructed by the outer products of 1-D vectors. 
At the outer product based composition step at each epoch of training, the number of parameters in the 3-D filters become the same as in the filters in standard CNNs, allowing a good gradient flow to flow throughout the network. This gradient flow also updates the parameters in the 1-D vectors, from which the 3-D filters are composed. At the next composition step, the weights in the 3-D filters are updated again, not by the gradient flow but by the outer product operation, to be projected onto the rank-1 subspace. By iterating this two-step update, all the 3-D filters in the network are trained to minimize the loss function while maintaining its rank-1 property. 
This is different from approaches which try to approximate the trained filters by low rank approximation
after the training has finished, e.g., like the low rank approximation in \cite{Jaderberg}. The composition operation is included in the training phase in our network, 
which directs the gradient flow in a different direction from that of standard CNNs, directing the solution to live on a rank-1 subspace.
In the testing phase, we do not need the outer product operation anymore, and can directly filter the input channels with the trained 1-D vectors treating them now as 1-D filters. That is, we take consecutive 1-D convolutions with the trained 1-D vectors, since the result is the same as being filtered with the 3-D filter constituted of the trained 1-D vectors. Therefore, the inference speed is exactly the same as that of the Flattened network. However, due to the better gradient flow, better parameters for the 1-D filters can be found with the proposed method, and more importantly, the network can be trained even in the case when the Flattened network can be not.\\
\indent We will also show that the convolution with rank-1 filters results in rank-deficient outputs, where the rank of the output is upper-bounded by a smaller bound than in normal CNNs. 
Therefore, the output feature vectors are constrained to live on a rank-deficient subspace in a high dimensional space. This coincides with the well-known belief that the feature vectors corresponding to images live on a low-dimensional manifold in a high dimensional space, and the fact that we get similar accuracy results with the rank-1 net can be another proof for this belief.\\
\indent We also explain in analogy to the bilateral-projection based 2-D principal component analysis(B2DPCA) what the 1-D vectors are trying to learn, and why the redundancy becomes reduced in the parameters with the rank-1 network. 
The reduction of the redundancy between the parameters is expressed by the reduced number of effective parameters, i.e., the number of parameters in the 1-D vectors. 
Therefore, the rank-1 net can be thought of as a compressed version of the standard CNN, and the reduced number of parameters as a smaller upper bound for the effective capacity of the standard CNN.
Compared with regularization methods, such as stochastic gradient descent, drop-out, and regularization methods, which do not reduce the excessive capacities of deep models as much as expected, the rank-1 projection reduces the capacity proportionally to the ratio of decrease in the number of parameters, and therefore, maybe can help to define a better upper bound for the effective capacity of deep networks. 

\section{Related Works}

The following works are related to our work. It is the work of the B2DPCA which gave us the insight for the rank-1 net. After we designed the rank-1 net, we found out that a similar research, i.e., the work on the Flattened network, has been done in the past. We explain both works below. 

\subsection{Bilateral-projection based 2DPCA}
In \cite{B2DPCA}, a bilateral-projection based 2D principal component analysis(B2DPCA) has been proposed, which minimizes the following energy functional:
\begin{equation} \label{bilateral}
[\P_{opt}, \Q_{opt}] = \AMIN_{\P, \Q} \| \X - \P\C\Q^T \|^2_F,
\end{equation}
where $\X \in R^{n \times m}$ is the two dimensional image, 
$\P \in R^{m \times l}$  and $\Q \in R^{n \times r}$ are the left- and right-
multiplying projection matrices, respectively, and $\C = \P^T\X\Q$ is the extracted feature matrix for the image $\X$. 
The optimal projection matrices $\P_{opt}$ and $\Q_{opt}$ are simultaneously constructed, where $\P_{opt}$ projects the column vectors of $\X$ to a subspace, 
while $\Q_{opt}$ projects the row vectors of $\X$ to another one.
To see why $\P$ is projecting the column vectors of $\X$ to a subspace, 
consider a simple example where $\P$ has $l$ column vectors:
\begin{equation} \label{P} 
\begin{array}{c} \P=
\left[
\begin{array}{cccc}
 |  &  |  &    &  |\\
 \p_1 & \p_2 & ... & \p_l\\
 |  &  |  &   & |\\
\end{array}
  \right],
 \end{array}
\end{equation} 
Then, left-multiplying $\P$ to the image $\X$, results in:
\begin{equation} \label{PX}
\begin{array}{c}
\P^T\X =\left[
\begin{array}{c}
\lrule \p_1^T \lrule \lrule\\
\lrule \p_2^T \lrule \lrule\\
      \vdots \\
\lrule \p_l^T \lrule \lrule\\
\end{array}
  \right]
\left[
\begin{array}{cccc}
 |  &  |  &   & |\\
 \x_{col_1} & \x_{col_2} & ... & \x_{col_m}\\
 |  &  |  &   & |\\
\end{array}
  \right] 
  \\ \\ =
  \left[
\begin{array}{cccc}
 \p_1^T\x_{col_1}  &  \p_1^T\x_{col_2}& ...  & \p_1^T\x_{col_m} \\
 \p_2^T\x_{col_1}  &  \p_2^T\x_{col_2}& ...  & \p_2^T\x_{col_m} \\
 \vdots  &  \vdots & \vdots  & \vdots \\
 \p_l^T\x_{col_1}  &  \p_l^T\x_{col_2}& ...  & \p_l^T\x_{col_m} \\
\end{array}
  \right],
 \end{array}
\end{equation}
where it can be observed that all the components in $\P^T\X$ are the projections of the column vectors of $\X$ onto the column vectors of $\P$.
Meanwhile, the right-multiplication of the matrix $\Q$ to $\X$ results in,
\begin{equation} \label{XQ} 
\begin{array}{c}
\X\Q = \left[
\begin{array}{cc}
\lrule \x_{row_1} \lrule  \lrule \\
\lrule \x_{row_2} \lrule  \lrule\\
   \vdots \\
\lrule \x_{row_n} \lrule  \lrule\\
\end{array}
  \right]
\left[
\begin{array}{cccc}
 |  &  |  &  & | \\
 \q_1 & \q_2 & ... & \q_r\\
 |  &  |  &   & |\\
\end{array}
  \right]
\\ \\ = 
\left[
\begin{array}{cccc}
\x_{row_1}\q_1  &  \x_{row_1}\q_2  & ... & \x_{row_1}\q_r \\
\x_{row_2}\q_1  &  \x_{row_2}\q_2  & ... & \x_{row_2}\q_r \\
 \vdots  & \vdots  & \vdots & \vdots \\
\x_{row_n}\q_1  &  \x_{row_n}\q_2  & ... & \x_{row_n}\q_r \\
\end{array}
  \right],
 \end{array}
\end{equation} 
where the components of $\X\Q$ are the projections of the row vectors of $\X$ onto the column vectors of $\Q$. 
From the above observation, we can see that the components of the feature matrix $\C = \P^T \X \Q \in R^{l \times r}$ is a result of simultaneously projecting the row vectors of $\X$ onto the column vectors of $\P$, and the column vectors of $\X$ onto the column vectors of $\Q$. It has been shown in \cite{B2DPCA}, that the advantage of the bilateral projection over the unilateral-projection scheme is that $\X$ can be represented effectively with smaller number of coefficients than in the unilateral case, i.e., a small-sized matrix $\C$ can well represent the image $\X$. This means that the bilateral-projection effectively removes the redundancies among both rows and columns of the image. 
Furthermore, since
\begin{equation} 
\begin{array}{c} 
\C = \P^T \X \Q
= 
\left[
\begin{array}{cccc}
\p^T_1 \X \q_1  &  \p^T_1 \X \q_2  &  ...
& \p^T_1 \X \q_r \\
\p^T_2 \X \q_1  &  \p^T_2 \X \q_2  &  ... & \p^T_2 \X \q_r \\
\vdots  &  \vdots   &  \vdots   & \vdots\\
\p^T_l \X \q_1  &  \p^T_l \X \q_2  & ... & \p^T_l \X \q_r \\
\end{array}\right] \\
= \left[
\begin{array}{cccc}
<\X, \p_1 \q^T_1>  &  <\X, \p_1 \q^T_2>  & ... &  <\X, \p_1 \q^T_r> \\
<\X, \p_2 \q^T_1>  &  <\X, \p_2 \q^T_2>  & ... & <\X, \p_2 \q^T_r> \\
\vdots & \vdots & \vdots  & \vdots \\
<\X, \p_l \q^T_1>  &  <\X, \p_l \q^T_2>  & ... & <\X, \p_l \q^T_r> \\
\end{array}\right], 
\end{array}
\end{equation}
it can be seen that the components of $\C$ are the 2-D projections of the image $\X$ onto the 2-D planes $\p_1 \q^T_1, \p_1 \q^T_2, ...\p_l \q^T_r$ made up by the outer products of the column vectors of $\P$ and $\Q$. The 2-D planes have a rank of one, since they are the outer products of two 1-D vectors. Therefore, the fact that $\X$ can be well represented by a small-sized $\C$ also implies the fact that $\X$ can be well represented by a few rank-1 2-D planes, i.e., only a few 
1-D vectors $\p_1, ...\p_l, \q_1, ....\q_r$, where $l << m$ and $r << n$.\\ 
\indent In the case of (\ref{bilateral}), the learned 2-D planes try to minimize the loss function 
\begin{equation}
 L= \| \X - \P\C\Q^T \|^2_F,
\end{equation}
i.e., try to learn to best approximate $\X$.
A natural question arises, if good rank-1 2-D planes can be obtained to minimize  other loss functions too, e.g., loss functions related to the image classification problem, such as
\begin{equation}
 L= \| y_{true} - y(\X,\P,\Q) \|^2_F,
\end{equation}
where $y_{true}$ denotes the true classification label for a certain input image $\X$, and  
$y(\X,\P,\Q)$ is the output of the network 
constituted by the outer products of the column vectors in the learned matrices $\P$ and $\Q$.
In this paper, we will show that it is possible to learn such rank-1 2-D planes, i.e., 2-D filters, if they are used in a deep structure. Furthermore, we extend the rank-1 2-D filter case to the rank-1 3-D filter case, where the rank-1 3-D filter is constituted as the outer product of three column vectors from three different learned matrices.

\subsection{Flattened Convolutional Neural Networks}

\begin{figure}
\centering
	\includegraphics[width=1\columnwidth]{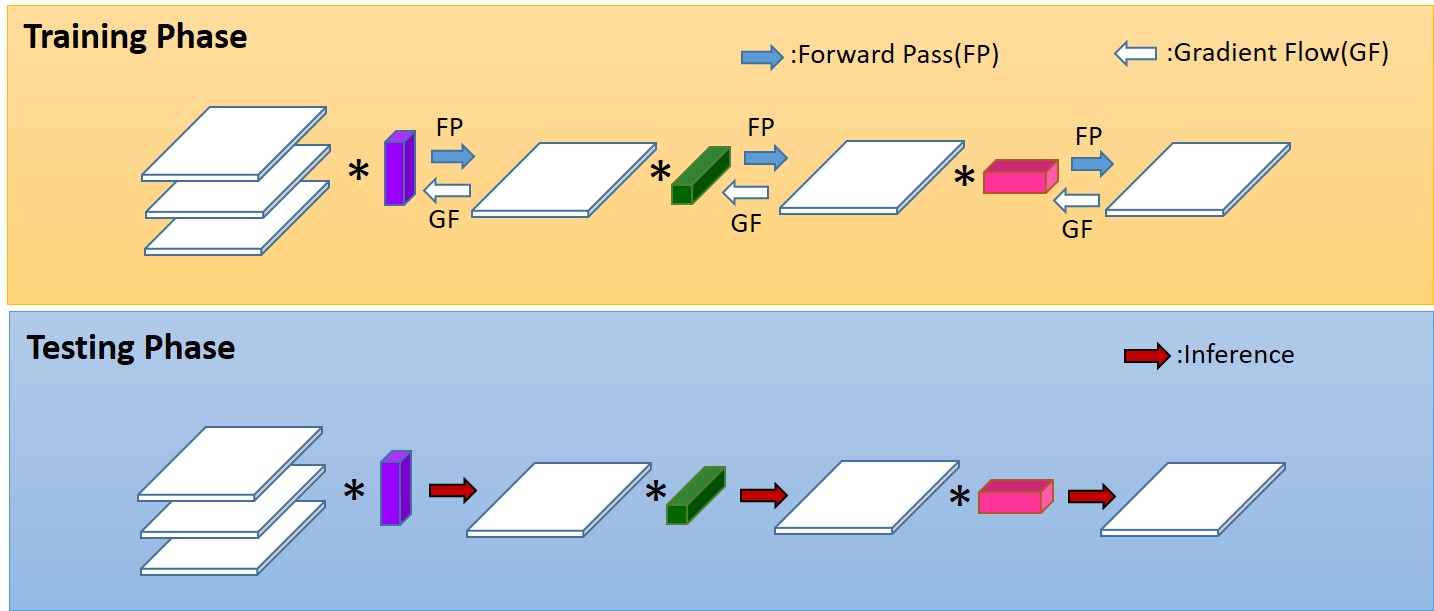} 
\caption{The structure in Flattened network. The same network structure of sequential use of 1-D filters is used in the training and testing phases.}
    \label{flattened-training}
\end{figure}

In \cite{Flattened}, the `Flattened CNN' has been proposed
for fast feed-forward execution by separating the conventional 3-D convolution filter
into three consecutive 1-D filters.
The 1-D filters sequentially convolve the input over different directions, i.e., the lateral, horizontal, and vertical directions. 
Figure \ref{flattened-training} shows the network structure of the Flattened CNN.
The Flattened CNN uses the same network structure in both the training and
the testing phases. This is in comparison with our proposed model, where we use a different network structure in the training phase as will be seen later. 
However, the consecutive use of 1-D filters in the training phase makes the training difficult. This is due to the fact that the gradient path becomes longer than in normal CNN, and therefore, the gradient flow vanishes faster while the error is more accumulating. 
Another reason is that the reduction in the number of parameters causes a gradient flow
different from that of the standard CNN, which is more difficult to find an
appropriate solution. This fact coincides with the experiments in \cite{Livni} which show that the gradient flow in
a network with small number of parameters cannot find good parameters.
Therefore, a particular weight initialization method has to be used with this setting. Furthermore, in \cite{Flattened}, the networks in the experiments have only three layers of convolution, which is maybe due to the fact of the difficulty in training networks with more layers.

\section{Proposed Method}

\begin{figure}
\centering
	\includegraphics[width=1\columnwidth]{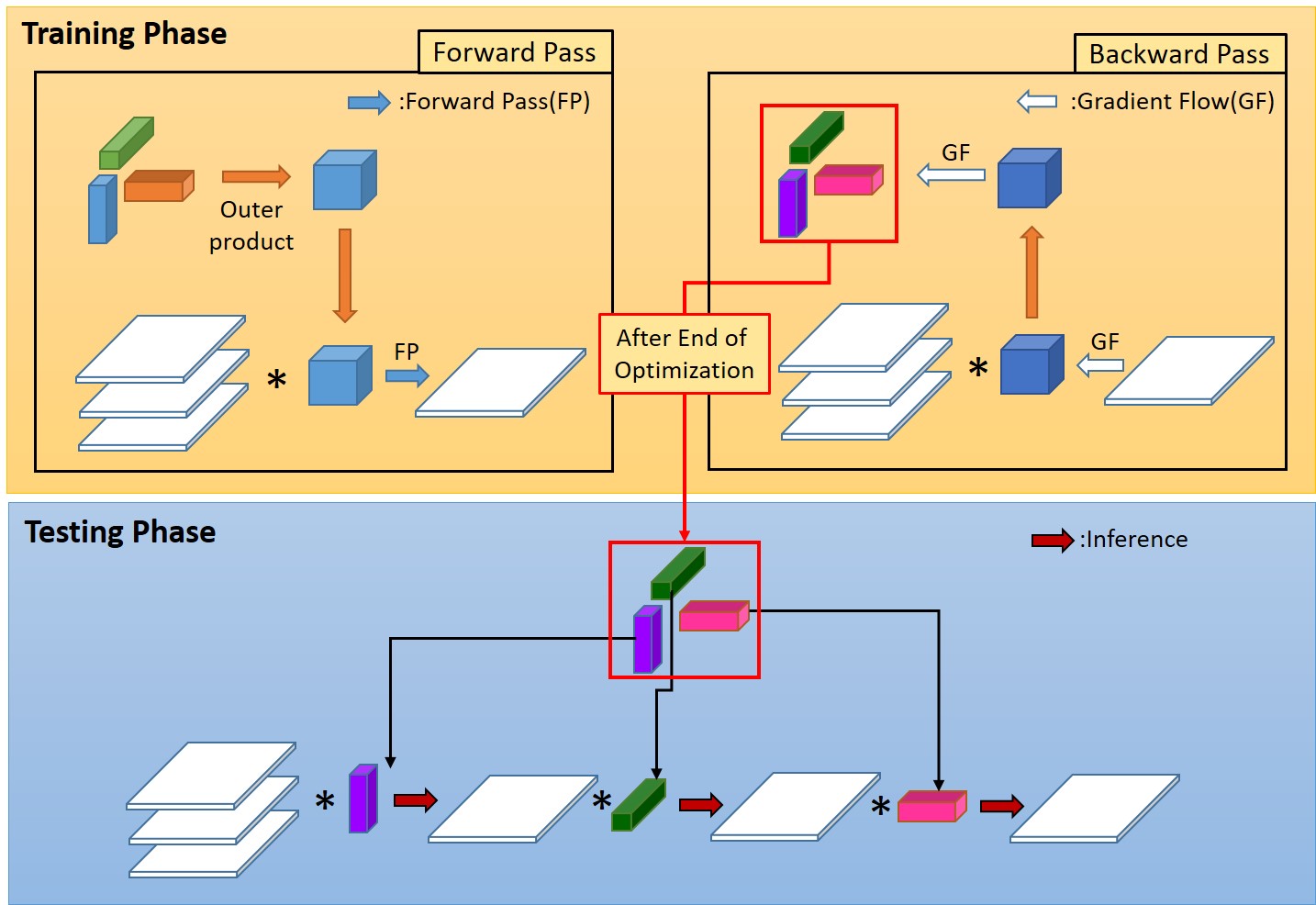} 
    	\caption{Proposed rank-1 neural network with different network structures in training and testing phases.}
    	    \label{proposed-training1}
\end{figure}

In comparison with other CNN models using 1-D rank-1 filters, we propose
the use of 3-D rank-1 filters($\w$) in the training stage, where the 3-D rank-1 filters are
constructed by the outer product of three 1-D vectors,
say $\p$, $\q$, and $\t$:
\begin{equation}
     \w = \p \otimes \q \otimes \t.
\end{equation}
This is an extension of the 2-D rank-1 planes used in the B2DPCA, where the 2-D planes
are constructed by $\w =  \p \otimes \q = \p\q^T$.
Figure \ref{proposed-training1} shows the training and the testing phases of the proposed method. The structure of the 
proposed network is different for the training phase and the testing phase.
In comparison with the Flattened network (Fig. \ref{flattened-training}), in the training phase, 
the gradient flow first flows through the 3-D rank-1 filters and then through the 1-D vectors.
Therefore, the gradient flow is different from that of the Flattened network resulting in a different and better solution of parameters in the 1-D vectors. 
The solution can be obtained even in large networks with the proposed method, 
for which the gradient flow in the Flattened network cannot obtain a solution at all.  
Furthermore, at test time, i.e., at the end of optimization, we can use the 1-D vectors directly as 1-D filters in the same manner as in the Flattened network, resulting in the same inference speed as the Flattened network(Fig. \ref{proposed-training1}). 

\begin{figure}
\centering
	\includegraphics[width=1\columnwidth]{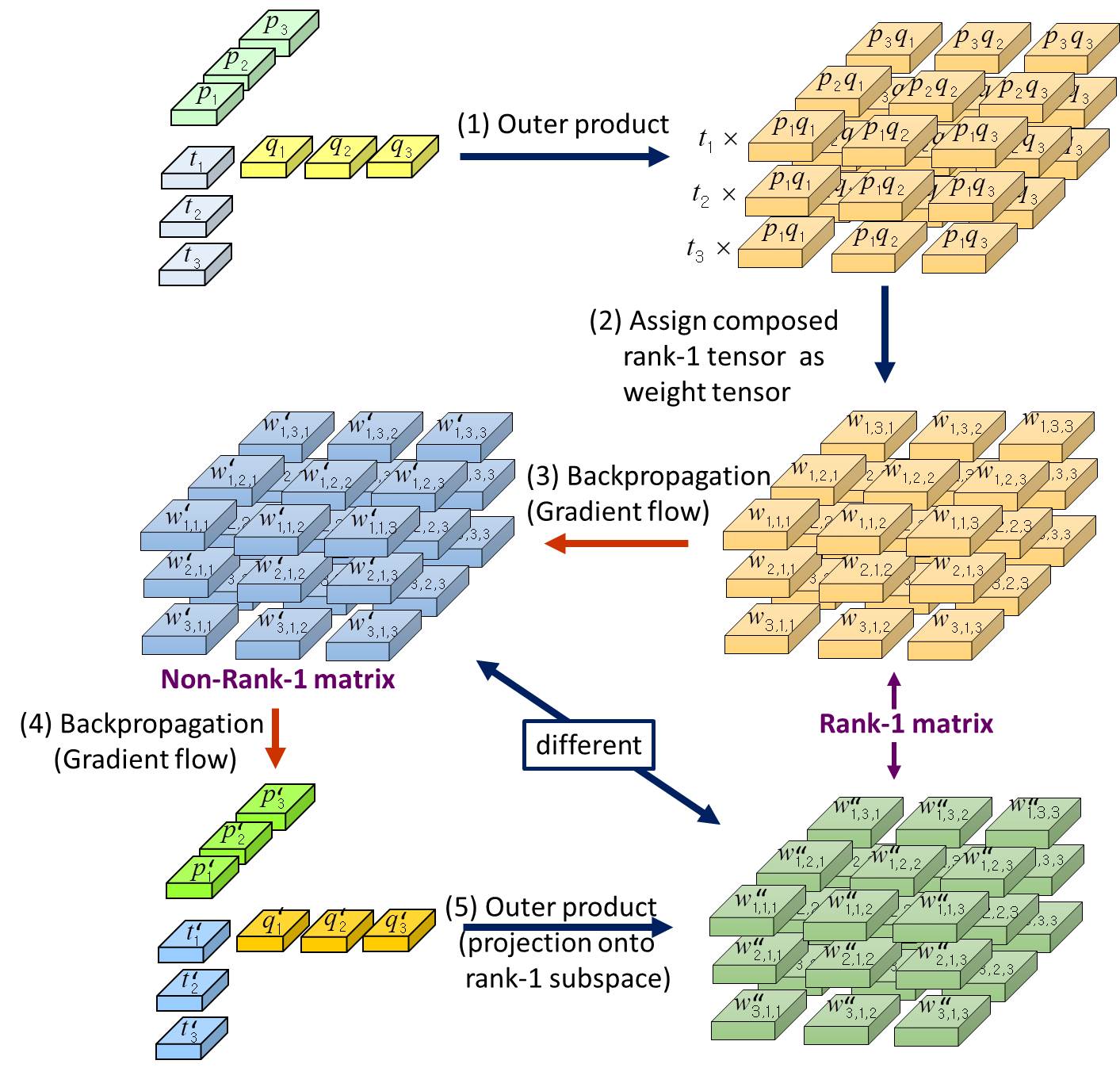} 
    	\caption{Steps in the training phase of the proposed rank-1 network.}
    	\label{proposed_training2}
\end{figure}

Figure \ref{proposed_training2} explains the training process with the proposed network structure in detail. At every epoch of the training phase, we first take the outer product of the three 1-D vectors $\p$, $\q$, and $\t$. Then, we assign the result of the outer product to the weight values of the 3-D convolution filter, i.e., for every weight value in the 3-D convolution filter $\w$, we assign 
\begin{equation} \label{FX}
  w_{i,j,k} = p_i q_j t_k, \,\, \forall_{i,j,k \in \Omega(\w)}  
\end{equation}
where, $i,j,k$ correspond to the 3-D coordinates in $\Omega(\w)$, the 3-D domain of the 3-D convolution filter $\w$. 
Since the matrix constructed by the outer product of vectors has always a rank of one, the 3-D convolution filter $\w$ is a rank-1 filter.\\ \indent During the back-propagation phase, every weight value in $\w$ will be updated by 
\begin{equation} \label{normal_update}
w'_{i,j,k} = w_{i,j,k} - \alpha \frac{\partial L}{\partial w_{i,j,k}},
\end{equation}
where $\frac{\partial L}{\partial w_{i,j,k}}$ denotes the gradient of the loss function $L$ with respect to the weight $w_{i,j,k}$, and $\alpha$ is the learning rate. 
In normal networks, $w'_{i,j,k}$ in (\ref{normal_update}) is the final updated weight value. However, the updated filter $\w'$ normally is not a rank-1 filter. This is due to the fact that the update in (\ref{normal_update}) is done in the direction which considers only the minimizing of the loss function and not the rank of the filter.\\
\indent With the proposed training network structure, we take a further update step, i.e., we update the 1-D vectors $\p$, $\q$, and $\t$: 
\begin{equation}
p'_{i} = p_{i} - \alpha \frac{\partial L}{\partial p_{i}}, \,\, \forall_{i \in \Omega(\p)}
\end{equation}
\begin{equation}
q'_{j} = q_{j} - \alpha \frac{\partial L}{\partial q_{j}}, \,\, \forall_{j \in \Omega(\q)}
\end{equation}
\begin{equation}
t'_{k} = t_{k} - \alpha \frac{\partial L}{\partial t_{k}}, \,\, \forall_{k \in \Omega(\t)}
\end{equation}
Here, $\frac{\partial L}{\partial p_{i}}$, $\frac{\partial L}{\partial q_{j}}$, and $\frac{\partial L}{\partial t_{k}}$ can be  calculated as

\begin{equation}
\frac{ \partial L }{ \partial p_i } = \sum_j \sum_k \frac{ \partial L}{ \partial w_{i,j,k} }\frac{ \partial w_{i,j,k}}{ \partial p_i}= \sum_j \sum_k \frac{ \partial L}{ \partial w_{i,j,k} }q_j t_k,
\end{equation}

\begin{equation}
\frac{ \partial L }{ \partial q_j } = \sum_i \sum_k \frac{ \partial L}{ \partial w_{i,j,k} }\frac{ \partial w_{i,j,k}}{ \partial q_j}= \sum_i \sum_k \frac{ \partial L}{ \partial w_{i,j,k} }p_i t_k,
\end{equation}

\begin{equation}
\frac{ \partial L }{ \partial t_k} = \sum_i \sum_j \frac{ \partial L}{ \partial w_{i,j,k} }\frac{ \partial w_{i,j,k}}{ \partial t_k}= \sum_i \sum_j \frac{ \partial L}{ \partial w_{i,j,k} }p_i q_j. 
\end{equation}

 At the next feed forward step of the back-propagation, an outer product of the updated 1-D vectors $\p$, $\q$, and $\t$ is taken to concatenate them back into the 3-D convolution filter $\w''$: 
\begin{equation} \label{next_update}
\begin{array}{ccc}
w''_{i,j,k} \!\!\!\!\!\!\!\!\! & = p'_{i}q'_{j}t'_{k}  = (p_{i} - \alpha \frac{\partial L}{\partial p_{i}})(q_{j} - \alpha \frac{\partial L}{\partial q_{j}})(t_{k} - \alpha \frac{\partial L}{\partial t_{k}})& \\
&\!\!\!\!\!\!\!\! = p_{i}q_{j}t_{k} - \alpha (p_{i}q_{j}\frac{\partial L}{\partial t_{k}}+q_{j}t_{k}\frac{\partial L}{\partial p_{i}} + p_{i}t_{k}\frac{\partial L}{\partial q_{j}})&\\
\!\!\!\!\!\!+ \!\!\!\!\!\!\!\!\! & {\alpha}^2 (p_{i}\frac{\partial L}{\partial q_{j}}\frac{\partial L}{\partial t_{k}}+ q_{j}\frac{\partial L}{\partial p_{i}}\frac{\partial L}{\partial t_{k}}+t_{k}\frac{\partial L}{\partial p_{i}}\frac{\partial L}{\partial t_{k}})-{\alpha}^3\frac{\partial L}{\partial p_{i}}\frac{\partial L}{\partial q_{j}}\frac{\partial L}{\partial t_{k}}& \\
 =&   w_{i,j,k} - \alpha \Delta_{i,j,k},  \,\, \forall_{i,j,k}, &\!\!\!\!\!\!\!\! \\
\end{array}
\end{equation} 
where
\begin{equation} \label{set}
\begin{array}{ccc}
\Delta_{i,j,k}\!\!\!\!\!\!\!\!\!\!\!\!\!\!\!\! & = p_{i}q_{j}\frac{\partial L}{\partial t_{k}}+q_{j}t_{k}\frac{\partial L}{\partial p_{i}} + p_{i}t_{k}\frac{\partial L}{\partial q_{j}}&\\
& \!\!\!\!\!\! - \alpha (p_{i}\frac{\partial L}{\partial q_{j}}\frac{\partial L}{\partial t_{k}}+ q_{j}\frac{\partial L}{\partial p_{i}}\frac{\partial L}{\partial t_{k}}+t_{k}\frac{\partial L}{\partial p_{i}}\frac{\partial L}{\partial t_{k}})+{\alpha}^2\frac{\partial L}{\partial p_{i}}\frac{\partial L}{\partial q_{j}}\frac{\partial L}{\partial t_{k}}.& \\
\end{array}
\end{equation}
As the outer product of 1-D vectors always results in a rank-1 filter, $\w''$ is a rank-1 filter as compared with $\w'$ which is not.
Comparing (\ref{normal_update}) with (\ref{next_update}), we get
\begin{equation}
w''_{i,j,k} = w'_{i,j,k} - \alpha (\Delta_{i,j,k}-\frac{\partial L}{\partial w_{i,j,k}}).
\end{equation}
Therefore, $\Delta_{i,j,k}-\frac{\partial L}{\partial w_{i,j,k}}$ is the incremental update vector which projects 
$\w'$ back onto the rank-1 subspace. 

\section{Property of rank-1 filters}

Below, we explain some properties of the 3-D rank-1 filters.

\subsection{Multilateral property of 3-D rank-1 filters}

We explain the bilateral property of the 2-D rank-1 filters in analogy to the B2DPCA.
The extension to the multilateral property of the 3-D rank-1 filters is then straightforward.
We first observe that a 2-D convolution can be seen as shifting inner products, where
each component $y(\r)$ at position $\r$ of the output matrix $\Y$ is computed as the inner product of a 2-D filter 
$\W$ and the image patch $\X(\r)$ centered at $\r$:
\begin{equation}
 y(\r) = <\W,\X(\r)>. 
\end{equation}
If $\W$ is a 2-D rank-1 filter, then,
\begin{equation}
 y(\r) = <\W,\X(\r)> = <\p\q^T, \X(\r)> = \p^T \X(\r) \q
\end{equation}
As has been explained in the case of B2DPCA, since $\p$ is multiplied to the rows of $\X(\r)$, $\p$ tries to extract the features from the rows of $\X(\r)$ which can minimize the loss function. That is, $\p$ searches the rows in all patches $\X(\r), \forall_{\r}$ for some common features which can reduce the loss function, while $\q$ looks for the features in the columns of the patches.
This is in analogy to the B2DPCA, where the bilateral projection removes the redundancies among the rows and columns in the 2-D filters.
Therefore, by easy extension, the 3-D rank-1 filters which are learned by the multilateral projection will have less redundancies among the rows, columns, and the channels than the normal 3-D filters in standard CNNs.

\subsection{Property of projecting onto a low dimensional subspace}

\begin{figure}
\centering
	\includegraphics[width=0.8\columnwidth]{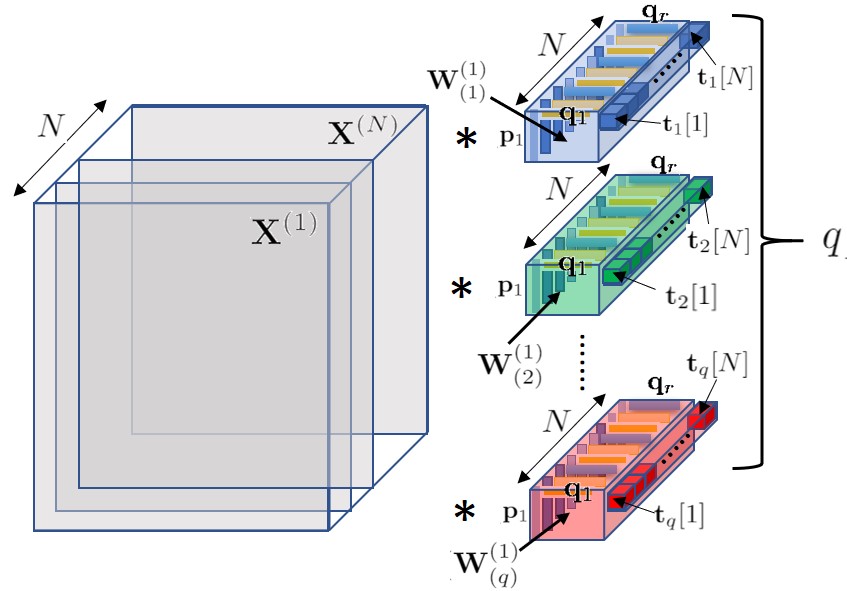}
    	\caption{Convolution filters of the proposed rank-1 network.}
    	    \label{HankelAnalysis}
\end{figure}

In this section, we show that the convolution with the rank-1 filters projects the output channels onto a low dimensional subspace. 
In \cite{DeepConvolution}, it has been shown via the block Hankel matrix formulation that the auto-reconstructing U-Net with insufficient number of filters results in a low-rank approximation of its input. Using the same block Hankel matrix formulation for the 3-D convolution, we can show that the 3-D rank-1 filter projects the input
onto a low dimensional subspace in a high dimension. 
To avoid confusion, we use the same definitions and notations as in \cite{DeepConvolution}. 
A wrap-around Hankel matrix $H_{d}(\f)$ of a function $\f = [f[1], f[2], \hdots ,f[n]]$ with respect to the number of columns $d$ is defined as 
\begin{equation}
H_{d}(\f) = 
\left[
\begin{array}{cccc}
f[1]  &  f[2]  & \hdots & f[d] \\
f[2]  &  f[3]  & \hdots & f[d+1]  \\
\vdots  &  \vdots  & \ddots & \vdots\\
f[n]  &  f[1]  & \hdots & f[d-1] \\
\end{array}
  \right] \in R^{n \times d}.
\end{equation}
Using the Hankel matrix, a convolution operation with a 1-D filter $\w$ of length $d$ can be expressed in a matrix-vector form as
\begin{equation}
\y = H_{d}(\f)\bar{\w},
\end{equation}
where $\bar{\w}$ is the flipped version of $\w$, and $\y$ is the output result of the convolution.\\
\indent The 2-D convolution can be expressed using the block Hankel matrix expression of the input channel. The block Hankel matrix of a 2-D input $\X = [\x_1, ..., \x_{n_2}] \in R^{n_1 \times n_2}$ with $\x_i \in R^{n_1}$ being the columns of $\X$, becomes 
\begin{equation}
H_{d_1,d_2}(\X) =  
\left[
\begin{array}{cccc}
H_{d_1} (\x_1)  &  H_{d_1} (\x_2)  & \hdots & H_{d_1} (\x_{d_2}) \\
H_{d_1} (\x_2)  &  H_{d_1} (\x_3)  & \hdots & H_{d_1} (\x_{d_2 +1}) \\
\vdots  &  \vdots  & \ddots & \vdots\\
H_{d_1} (\x_{n_2})  &  H_{d_1} (\x_1)  & \hdots & H_{d_1} (\x_{d_2 -1}) \\
\end{array}
  \right], 
\end{equation}
where $H_{d_1,d_2}(\X) \in R^{n_1 n_2 \times d_1 d_2}$ and $H_{d_1}(\x_{i}) \in R^{n_1 \times d_1}$.
With the block Hankel matrix, a single-input single-output 2-D convolution with a 2-D filter $\W$ of size $d_1 \times d_2$ can be expressed in matrix-vector form,  
\begin{equation}
VEC(\Y) = H_{d_1,d_2}(\X) VEC(\W),
\end{equation}
where $VEC(\Y)$ denotes the vectorization operation by stacking up the column vectors of the 2-D matrix $\Y$.\\
\indent In the case of multiple input channels $\X^{(1)} \hdots  \X^{(N)}$, the block Hankel matrix is extended to
\begin{equation}
H_{d_1,d_2 | N}\left( [\X^{(1)} \hdots  \X^{(N)}]  \right)= \left[ H_{d_1,d_2}(\X^{(1)}) \hdots H_{d_1,d_2}(\X^{(N)})\right], 
\end{equation}
and a single output of the multi-input convolution with multiple filters becomes
\begin{equation}
VEC(\Y^{(i)})=\sum_{j=1}^{N}H_{d_1,d_2}(\X^{(j)})VEC(\W_{(i)}^{(j)}), \,\,\, i=1,\hdots,q,
\end{equation}
where $q$ is the number of filters.
Last, the matrix-vector form of the multi-input multi-output convolution resulting in multiple outputs $\Y^{(1)} \hdots \Y^{(q)}$ can be expressed as
\begin{equation}
\Y = H_{d_1,d_2 | N}\left( [\X^{(1)} \hdots  \X^{(N)}]  \right) \W,
\end{equation}
where
\begin{equation}
\Y = [VEC(\Y^{(1)})  \, \hdots \, VEC(\Y^{(q)})]
\end{equation}
and
\begin{equation}
\W =  
\left[
\begin{array}{ccc}
VEC(\W_{(1)}^{(1)})  &  \hdots  &  VEC(\W_{(q)}^{(1)}) \\
\vdots  &  \ddots  &  \vdots \\
VEC(\W_{(1)}^{(N)})  &  \hdots  &  VEC(\W_{(q)}^{(N)}) \\
\end{array}
  \right].
\end{equation}
To calculate the upper bound of the rank of $\Y$, we use the rank inequality 
\begin{equation}
rank(\mathbf{AB}) \leq  min\{rank(\mathbf{A}),rank(\mathbf{B})\}
\end{equation}
on $\Y$ to get
\begin{equation}
rank(\Y)\!\leq\!min \{rank H_{d_1,d_2|N}\!\left(\![\X^{(1)}\hdots\X^{(N)}]\!\right)\!,rank(\W)\}.
\end{equation}
Now to investigate the rank of $\W$, we first observe that 
\begin{equation}
\W =  
\left[
\begin{array}{ccc}
\t_1[1]
 VEC(\p_1 \otimes  \q_1 )  &  \hdots  &  \t_q[1]
VEC(\p_1 \otimes  \q_1 )  \\
\vdots  &  \ddots  &  \vdots \\
\t_1[N]
VEC(\p_l \otimes  \q_r )  &  \hdots  &  \t_q[N]
VEC(\p_l \otimes  \q_r ) \\
\end{array}
  \right]
\end{equation}
as can be seen in Fig. \ref{HankelAnalysis}.\\
\indent  Then, expressing $\W$ as the stack of its sub-matrices,
\begin{equation}
\W =  
\left[
\begin{array}{ccc}
\W_1 \\
\vdots\\
\W_s \\
\vdots\\
\W_N \\
\end{array}
  \right] \in R^{Nd_1d_2 \times q},
\end{equation}
where
\begin{equation}
\W_s = 
\left[
\begin{array}{ccc}
\t_1[s]VEC(\p_i\otimes\q_j) & \hdots & \t_q[s]VEC(\p_i\otimes\q_j)\\
 \end{array}
\right],
\end{equation}
which columns are the vectorized forms of the 2-D slices in the 3-D filters which convolve with the $s$-th image. We observe that all the sub-matrices $\W_s \in R^{d_1d_2  \times  q}, (s=1,...N)$ have a rank of 1, since all the column vectors in $\W_s$ are in the same direction and differ only in their magnitudes, i.e., by the different values of $\t_1[s], ..., \t_q[s]$.
Therefore, the upper bound of $rank(\W)$ is $min\{N,q\}$ instead of $min\{Nd_1d_2,q\}$ which is the upper bound we get if we use non-rank-1 filters.\\
\indent As a result, the output $\Y$ is upper bounded as
\begin{equation}
rank(\Y) \leq a,
\end{equation}
where
\begin{equation}
\label{ranka}
\begin{array}{cc}
a = min \{rank H_{d_1,d_2|N}\left([\X^{(1)}\hdots\X^{(N)}]\right),\\
\mbox{number of input channels ($N$)},\\
\mbox{number of filters ($q$)}\}.
\end{array}
\end{equation}
As can be seen from (\ref{ranka}), the upper bound is determined by the ranks of Hankel matrices of the input channels or the numbers of input channels or filters.
In common deep neural network structures, the number of filters are normally larger than the number of input channels, e.g., the VGG-16 uses in every layer a number of filters larger or equal to the number of input channels. So if we use the same structure for the proposed rank-1 network as in the VGG-16 model, the upper bound will be determined mainly by the number of input channels. 
Therefore, the outputs of layers in the proposed CNN are constrained to live on sub-spaces having lower ranks than the sub-spaces on which the outputs of layers in standard CNNs live. Since the output of a certain layer becomes the input of the next layer, the difference in the rank between the standard and the proposed rank-1 CNN accumulates in higher layers. Therefore, the final output of the proposed rank-1 CNN lives on a sub-space of much lower rank than the output of the standard CNN. 

\section{Experiments}

\begin{figure}
\centering
	\includegraphics[width=1\columnwidth]{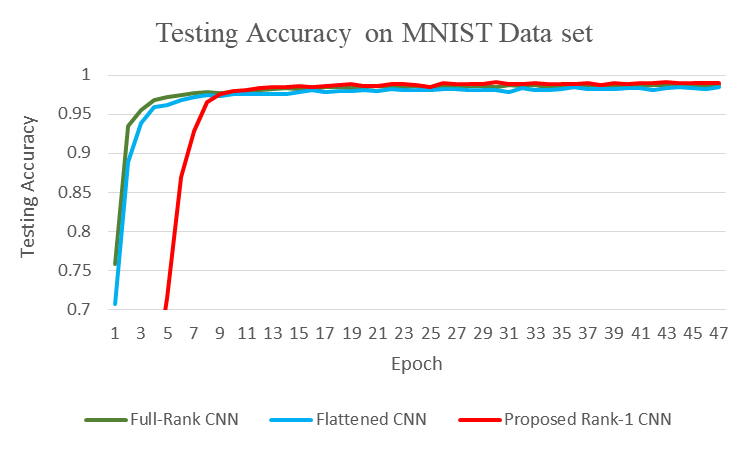} 
    	\caption{Comparison of test accuracy on the MNIST dataset.}
    	\label{MNIST}
\end{figure}

\begin{figure}
\centering
	\includegraphics[width=1\columnwidth]{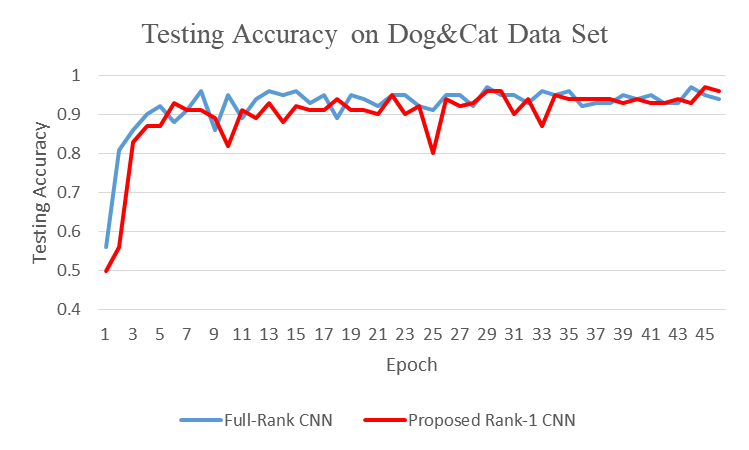} 
    	\caption{Comparison of test accuracy on the `Dog and Cat' dataset.}
    	\label{DogAndCat}
\end{figure}

We compared the performance of the proposed model with the standard CNN and the Flattened CNN model \cite{Flattened}.
We used the same number of layers for all the models, where for the Flattened CNN we regarded the combination of the lateral, vertical, and horizontal 1-D convolutional layers as a single layer. Furthermore, we used the same numbers of input and output channels in each layer for all the models, and also the same ReLU, Batch normalization, and dropout operations.
The codes for the proposed rank-1 CNN will be opened at https://github.com/petrasuk/Rank-1-CNN.\\
\indent Table 1-3 show the different structures of the models used for each dataset in the training stage. 
The outer product operation of three 1-D filters $\p$, $\q$, and $\t$ into a 3-D rank-1 filter $\w$ is denoted as $\w \doteq \p \otimes \q \otimes \t$ in the tables.
The datasets that we used in the experiments are the MNIST, the CIFAR10, and the `Dog and Cat'(https://www.kaggle.com/c/dogs-vs-cats) datasets. 
We used different structures for different datasets. 
For the experiments on the MNIST and the CIFAR10 datasets, we trained on 50,000 images, and 
then tested on 100 batches each consisting of 100 random images, and calculated the overall average accuracy. The sizes of the images in the MNIST and the CIFAR10 datasets are $28 \times 28$ and $32 \times 32$, respectively.
For the `Dog and Cat' dataset, we trained on 24,900 training images (size $224 \times 224$), and tested on a set of 100 test images.\\ 
\indent The proposed rank-1 CNN achieved a slightly larger testing accuracy on the MNIST dataset than the other two models (Fig. \ref{MNIST}). This is maybe due to the fact that the MNIST dataset is in its nature a low-ranked one, for which the proposed method can find the best approximation since the proposed method constrains the solution to a low rank sub-space. With the CIFAR10 dataset, the accuracy is slightly less than that of the standard CNN which maybe due to the fact that the images in the CIFAR10 datasets are of higher ranks than those in the MNIST dataset. 
However, the testing accuracy of the proposed CNN is higher than that of the Flattened CNN which shows the fact that the better gradient flow in the proposed CNN model achieves a better solution. The `Dog and Cat' dataset was used in the experiments to verify the performance of the proposed CNN on real-sized images and on a
deep structure. In this case, we could not train the Flattened network due to memory issues. We used the Tensorflow API, and somehow, the Tensorflow API requires much more GPU memory for the Flattened network than the proposed rank-1 network. 
We also believe that, even if there is no memory issue, with this deep structure, the Flattened network cannot find good parameters at all due to the limit of the bad gradient flow in the deep structure. 
The Standard CNN and the proposed CNN achieved similar test accuracy as can be seen in Fig. \ref{DogAndCat}.

\begin{table}[h]
  \begin{center}
    \caption{Structure of CNN for MNIST dataset}
    \begin{tabular}{c|c|c}
        \textbf{Standard CNN} & \textbf{Flattened CNN} & \textbf{Proposed CNN}\\
         \hline
     \multicolumn{3}{c}{Conv1: 64 filters, each filter constituted as:}\\
          \hline     
      \multirow{4}{*}{$1\times3\times3$ conv} & $1\times1\times1\times$ conv &$\w_{1} \doteq \p_{1}(1\times3\times1)$\\ 
         & $1\times3\times1$ conv & $\otimes\q_{1}(1\times1\times3)$\\ 
      & $1\times1\times3$ conv & $\otimes\t_{1}(1\times1\times1)$\\ 
          &    &  $1\times3\times3$ conv     \\ 
       \hline
 \multicolumn{3}{c}{Conv2: 64 filters, each filter constituted as:}\\
          \hline     
      \multirow{4}{*}{$64\times3\times3$ conv} & $64\times1\times1$ conv &$\w_{2} \doteq \p_{2}(1\times3\times1)$\\ 
         & $1\times3\times1$ conv & $\otimes\q_{2}(1\times1\times3)$\\ 
              & $1\times1\times3$ conv & $\otimes\t_{2}(64\times1\times1)$     \\ 
              &       & $64\times3\times3$ conv\\ 
       \hline
     \multicolumn{3}{c}{Max Pool ($\frac{1}{2})$}\\
        \hline
     \multicolumn{3}{c}{Conv3: 144 filters, each filter constituted as:}\\
          \hline     
      \multirow{4}{*}{$64\times3\times3$ conv} & $64\times1\times1$ conv &$\w_{3} \doteq \p_{3}(1\times3\times1)$\\ 
         & $1\times3\times1$ conv & $\otimes\q_{3}(1\times1\times3)$\\ 
        & $1\times1\times3$ conv & $\otimes\t_{3}(64\times1\times1)$     \\ 
       &       & $64\times3\times3$ conv\\ 
       \hline
     \multicolumn{3}{c}{Conv4: 144 filters, each filter constituted as:}\\
          \hline     
  \multirow{4}{*}{$144\times3\times3$ conv} & $144\times1\times1$ conv &$\w_{4} \doteq \p_{4}(1\times3\times1)$\\ 
         & $1\times3\times1$ conv & $\otimes\q_{4}(1\times1\times3)$\\ 
        & $1\times1\times3$ conv & $\otimes\t_{4}(144\times1\times1)$     \\ 
       &       & $144\times3\times3$ conv\\ 
   \hline
     \multicolumn{3}{c}{Max Pool ($\frac{1}{2})$}\\
       \hline
     \multicolumn{3}{c}{Conv5: 144 filters, each filter constituted as:}\\
          \hline
\multirow{4}{*}{$144\times3\times3$ conv} & $144\times1\times1$ conv &$\w_{5} \doteq \p_{5}(1\times3\times1)$\\ 
         & $1\times3\times1$ conv & $\otimes\q_{5}(1\times1\times3)$\\ 
        & $1\times1\times3$ conv & $\otimes\t_{5}(144\times1\times1)$     \\ 
       &       & $144\times3\times3$ conv\\ 
   \hline
     \multicolumn{3}{c}{Conv6: 256 filters, each filter constituted as:}\\
          \hline
          \multirow{4}{*}{$144\times3\times3$ conv} & $144\times1\times1$ conv &$\w_{6} \doteq \p_{6}(1\times3\times1)$\\ 
         & $1\times3\times1$ conv & $\otimes\q_{6}(1\times1\times3)$\\ 
        & $1\times1\times3$ conv & $\otimes\t_{6}(144\times1\times1)$     \\ 
       &       & $144\times3\times3$ conv\\ 
   \hline
     \multicolumn{3}{c}{Conv7: 256 filters, each filter constituted as:}\\
          \hline   
   \multirow{4}{*}{$256\times3\times3$ conv} & $256\times1\times1$ conv &$\w_{7} \doteq \p_{7}(1\times3\times1)$\\ 
         & $1\times3\times1$ conv & $\otimes\q_{7}(1\times1\times3)$\\ 
        & $1\times1\times3$ conv & $\otimes\t_{7}(256\times1\times1)$     \\ 
       &       & $256\times3\times3$ conv\\ 
  \hline
       \multicolumn{3}{c}{FC 2048 + Batch Normalization + ReLU + Drop Out (Prob. = 0.5)}\\
       \hline
     \multicolumn{3}{c}{FC 1024 + Batch Normalization + ReLU + Drop Out (Prob. = 0.5)}\\
       \hline     
     \multicolumn{3}{c}{FC 10 + ReLU + Drop Out (Prob. = 0.5)}\\
       \hline     
     \multicolumn{3}{c}{Soft-Max}\\
       \hline     
    \end{tabular}
  \end{center}
      \label{tab:table2}
\end{table}

\begin{table}[h!]
  \label{tab:table1}
  \begin{center}
    \caption{Structure of CNN for CIFAR-10 dataset}
    \begin{tabular}{c|c|c}
        \textbf{Standard CNN} & \textbf{Flattened CNN} & \textbf{Proposed CNN}\\
      \hline
     \multicolumn{3}{c}{Conv1: 64 filters, each filter constituted as:}\\
             \hline     
      \multirow{4}{*}{$3\times3\times3$ conv} & $3\times1\times1$ conv &$\w_{1} \doteq \p_{1}(1\times3\times1)$\\ 
         & $1\times3\times1$ conv & $\otimes\q_{1}(1\times1\times3)$\\ 
      & $1\times1\times3$ conv & $\otimes\t_{1}(3\times1\times1)$\\ 
          &    &  $3\times3\times3$ conv     \\ 
      \hline
      \multicolumn{3}{c}{ReLU + Batch Normalization}\\
       \hline
    \multicolumn{3}{c}{Conv2: 64 filters, each filter constituted as:}\\
             \hline     
      \multirow{4}{*}{$64\times3\times3$ conv} & $64\times1\times1$ conv &$\w_{2} \doteq \p_{2}(1\times3\times1)$\\ 
         & $1\times3\times1$ conv & $\otimes\q_{2}(1\times1\times3)$\\ 
              & $1\times1\times3$ conv & $\otimes\t_{2}(64\times1\times1)$     \\ 
              &       & $64\times3\times3$ conv\\ 
       \hline
     \multicolumn{3}{c}{ReLU + Max Pool ($\frac{1}{2})$ + Drop Out (Prob. = 0.5)}\\
            \hline
        \multicolumn{3}{c}{Conv3: 144 filters, each filter constituted as:}\\
             \hline     
      \multirow{4}{*}{$64\times3\times3$ conv} & $64\times1\times1$ conv &$\w_{3} \doteq \p_{3}(1\times3\times1)$\\ 
         & $1\times3\times1$ conv & $\otimes\q_{3}(1\times1\times3)$\\ 
        & $1\times1\times3$ conv & $\otimes\t_{3}(64\times1\times1)$     \\ 
       &       & $64\times3\times3$ conv\\ 
       \hline
     \multicolumn{3}{c}{ReLU + Batch Normalization}\\
       \hline
    \multicolumn{3}{c}{Conv4: 144 filters, each filter constituted as:}\\
          \hline     
  \multirow{4}{*}{$144\times3\times3$ conv} & $144\times1\times1$ conv &$\w_{4} \doteq \p_{4}(1\times3\times1)$\\ 
         & $1\times3\times1$ conv & $\otimes\q_{4}(1\times1\times3)$\\ 
        & $1\times1\times3$ conv & $\otimes\t_{4}(144\times1\times1)$     \\ 
       &       & $144\times3\times3$ conv\\ 
   \hline
      \multicolumn{3}{c}{ReLU + Max Pool ($\frac{1}{2})$ +Drop Out (Prob. = 0.5)}\\
       \hline
    \multicolumn{3}{c}{Conv5: 256 filters, each filter constituted as:}\\
          \hline     
    \multirow{4}{*}{$144\times3\times3$ conv} & $144\times1\times1$ conv &$\w_{5} \doteq \p_{5}(1\times3\times1)$\\ 
         & $1\times3\times1$ conv & $\otimes\q_{5}(1\times1\times3)$\\ 
        & $1\times1\times3$ conv & $\otimes\t_{5}(144\times1\times1)$     \\ 
       &       & $144\times3\times3$ conv\\ 
   \hline
      \multicolumn{3}{c}{ReLU + Batch Normalization}\\
       \hline
    \multicolumn{3}{c}{Conv6: 256 filters, each filter constituted as:}\\
          \hline     
   \multirow{4}{*}{$256\times3\times3$ conv} & $256\times1\times1$ conv &$\w_{6} \doteq \p_{6}(1\times3\times1)$\\ 
         & $1\times3\times1$ conv & $\otimes\q_{6}(1\times1\times3)$\\ 
        & $1\times1\times3$ conv & $\otimes\t_{6}(256\times1\times1)$     \\ 
       &       & $256\times3\times3$ conv\\ 
   \hline
       \multicolumn{3}{c}{ReLU + Max Pool ($\frac{1}{2})$ + Drop Out (Prob. = 0.5)}\\
       \hline
     \multicolumn{3}{c}{FC 1024 + Batch Normalization + ReLU + Drop Out (Prob. = 0.5)}\\
       \hline     
     \multicolumn{3}{c}{FC 512 + Batch Normalization + ReLU + Drop Out (Prob. = 0.5)}\\
       \hline     
     \multicolumn{3}{c}{FC 10}\\
       \hline     
     \multicolumn{3}{c}{Soft-Max}\\
       \hline     
    \end{tabular}
  \end{center}
\end{table}

\begin{table}[h] 
  \begin{center}
    \caption{Structure of CNN for `Dog and Cat' dataset}
                \label{tab:table3}
    \begin{tabular}{c|c}
        \textbf{Standard CNN} &  \textbf{Proposed CNN}\\
         \hline
     \multicolumn{2}{c}{Conv1: 64 filters, each filter constituted as:}\\
          \hline
      \multirow{3}{*}{$3\times3\times3$ conv} & $\w_{1} \doteq \p_{1}(1\times3\times1) \otimes$\\ 
             & $\q_{1}(1\times1\times3)\otimes\t_{1}(3\times1\times1)$\\ 
              &  $3\times3\times3$ conv     \\ 
       \hline
     \multicolumn{2}{c}{Conv2: 64 filters, each filter constituted as:}\\
          \hline       
\multirow{3}{*}{$64\times3\times3$ conv} & $\w_{2} \doteq \p_{2}(1\times3\times1) \otimes$\\ 
             & $\q_{2}(1\times1\times3)\otimes\t_{2}(64\times1\times1)$\\ 
              &  $64\times3\times3$ conv     \\ 
 \hline
     \multicolumn{2}{c}{Batch Normalization + ReLU + Max Pool ($\frac{1}{2})$}\\
        \hline
     \multicolumn{2}{c}{Conv3: 144 filters, each filter constituted as:}\\
          \hline        
 \multirow{3}{*}{$64\times3\times3$ conv} & $\w_{3} \doteq \p_{3}(1\times3\times1) \otimes$\\ 
             & $\q_{3}(1\times3\times1)\otimes\t_{3}(64\times1\times1)$\\ 
              &  $64\times3\times3$ conv     \\ 
       \hline
     \multicolumn{2}{c}{ReLU}\\
     \hline
     \multicolumn{2}{c}{Conv4: 144 filters, each filter constituted as:}\\
          \hline     
 \multirow{3}{*}{$144\times3\times3$ conv} & $\w_{4} \doteq \p_{4}(1\times3\times1) \otimes$\\ 
             & $\q_{4}(1\times1\times3)\otimes\t_{4}(144\times1\times1)$\\ 
              &  $144\times3\times3$ conv     \\ 
       \hline
    \multicolumn{2}{c}{Batch Normalization + ReLU + Max Pool ($\frac{1}{2})$}\\
       \hline
     \multicolumn{2}{c}{Conv5: 256 filters, each filter constituted as:}\\
          \hline       
   \multirow{3}{*}{$144\times3\times3$ conv} & $\w_{5} \doteq \p_{5}(1\times3\times1) \otimes$\\ 
         & $\q_{5}(1\times1\times3)\otimes\t_{5}(144\times1\times1)$\\ 
         &  $144\times3\times3$ conv     \\ 
       \hline
    \multicolumn{2}{c}{ReLU}\\
          \hline
   \multicolumn{2}{c}{Conv6: 256 filters, each filter constituted as:}\\
          \hline
   \multirow{3}{*}{$256\times3\times3$ conv} & $\w_{6} \doteq \p_{6}(1\times3\times1) \otimes$\\ 
         & $\q_{6}(1\times1\times3)\otimes\t_{6}(256\times1\times1)$\\ 
         &  $256\times3\times3$ conv     \\ 
       \hline
    \multicolumn{2}{c}{Batch Normalization + ReLU + Max Pool ($\frac{1}{2})$}\\
       \hline
     \multicolumn{2}{c}{Conv7: 256 filters, each filter constituted as:}\\
          \hline       
   \multirow{3}{*}{$256\times3\times3$ conv} & $\w_{7} \doteq \p_{7}(1\times3\times1) \otimes$\\ 
         & $\q_{7}(1\times1\times3)\otimes\t_{7}(256\times1\times1)$\\ 
         &  $256\times3\times3$ conv     \\ 
       \hline
    \multicolumn{2}{c}{ReLU}\\
          \hline
     \multicolumn{2}{c}{Conv8: 484 filters, each filter constituted as:}\\
          \hline
   \multirow{3}{*}{$256\times3\times3$ conv} & $\w_{8} \doteq \p_{8}(1\times3\times1) \otimes$\\ 
         & $\q_{8}(1\times1\times3)\otimes\t_{8}(256\times1\times1)$\\ 
         &  $256\times3\times3$ conv     \\ 
       \hline
    \multicolumn{2}{c}{ReLU}\\
             \hline
     \multicolumn{2}{c}{Conv9: 484 filters, each filter constituted as:}\\
          \hline
   \multirow{3}{*}{$484\times3\times3$ conv} & $\w_{9} \doteq \p_{9}(1\times3\times1) \otimes$\\ 
         & $\q_{9}(1\times1\times3)\otimes\t_{9}(484\times1\times1)$\\ 
         &  $484\times3\times3$ conv     \\ 
       \hline
    \multicolumn{2}{c}{Batch Normalization + ReLU + Max Pool ($\frac{1}{2})$}\\
          \hline
     \multicolumn{2}{c}{Conv10: 484 filters, each filter constituted as:}\\
          \hline
   \multirow{3}{*}{$484\times3\times3$ conv} & $\w_{10} \doteq \p_{10}(1\times3\times1) \otimes$\\ 
         & $\q_{10}(1\times1\times3)\otimes\t_{10}(484\times1\times1)$\\ 
         &  $484\times3\times3$ conv     \\ 
       \hline
    \multicolumn{2}{c}{ReLU}\\
                \hline
 \multicolumn{2}{c}{Conv11: 484 filters, each filter constituted as:}\\
          \hline
   \multirow{3}{*}{$484\times3\times3$ conv} & $\w_{11} \doteq \p_{11}(1\times3\times1) \otimes$\\ 
         & $\q_{11}(1\times1\times3)\otimes\t_{11}(484\times1\times1)$\\ 
         &  $484\times3\times3$ conv     \\ 
       \hline
    \multicolumn{2}{c}{Batch Normalization + ReLU + Max Pool ($\frac{1}{2})$}\\
       \hline
     \multicolumn{2}{c}{FC 1024 + Batch Normalization + ReLU}\\
       \hline     
     \multicolumn{2}{c}{FC 512 + Batch Normalization + ReLU}\\       
     \hline     
     \multicolumn{2}{c}{FC 2}\\
 \hline     
     \multicolumn{2}{c}{Soft-Max}\\
       \hline     
    \end{tabular}
  \end{center}
\end{table}

\begin{figure}
\centering
	\includegraphics[width=1\columnwidth]{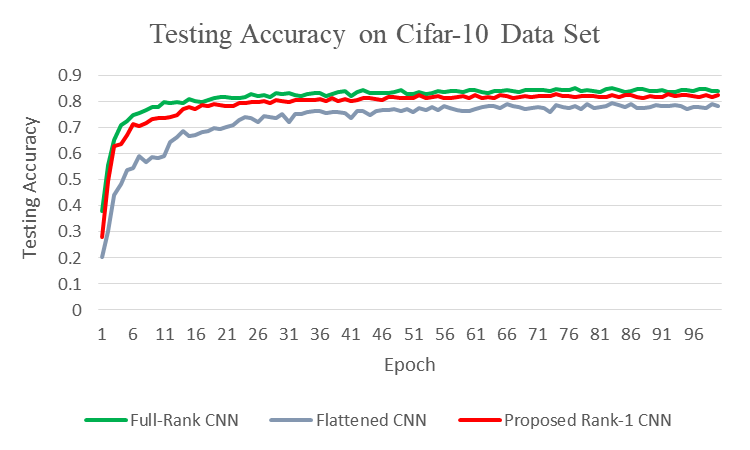} 
    	\caption{Comparison of test accuracy on the CIFAR10 dataset.}
    	\label{CIFAR10}
\end{figure}

\ifCLASSOPTIONcaptionsoff
  \newpage
\fi

%








\end{document}